\documentclass[letterpaper, 10 pt, journal, twoside]{ieeetran}
\usepackage{amsmath,amsfonts, amssymb}
\usepackage{algorithmic}
\usepackage{hyperref}
\usepackage{algorithm}
\usepackage{array}
\usepackage[caption=false,font=footnotesize,labelfont=sf,textfont=sf]{subfig}
\usepackage{textcomp}
\usepackage{stfloats}
\usepackage{url}
\usepackage{verbatim}
\usepackage{graphicx}
\usepackage{cite}
\usepackage{adjustbox}
\usepackage{booktabs}
\usepackage{threeparttable}
\usepackage{tabularx}
\usepackage{xcolor}
\usepackage{soul}

\definecolor{highlightcolor}{RGB}{255, 255, 0}
\sethlcolor{highlightcolor}

\begin{document}

\title{LIO-GVM: an Accurate, Tightly-Coupled Lidar-Inertial Odometry with Gaussian Voxel Map }

\author{Xingyu Ji$^{1}$ , ShengHai Yuan$^{1}$, Pengyu Yin$^{1}$, Lihua Xie$^{1}$
\thanks{Manuscript received: August, 04, 2023; Revised October, 22, 2023; Accepted December, 21, 2023.}
\thanks{This paper was recommended for publication by Editor Javier Civera upon evaluation of the Associate Editor and Reviewers’ comments.
This work was supported by the National Research Foundation, Singapore under its Medium Sized Center for Advanced Robotics Technology Innovation.} 
\thanks{$^{1}$ Authors are with the Centre for Advanced Robotics Technology Innovation (CARTIN), School of Electrical and Electronic Engineering, Nanyang Technological University, Singapore.{\tt\footnotesize \{xingyu001, shyuan, pengyu001, elhxie\}@ntu.edu.sg}}%

\thanks{Digital Object Identifier (DOI): see top of this page.}
}




\maketitle

\markboth{IEEE Robotics and Automation Letters. Preprint Version. Accepted Jan, 2024}
{Ji \MakeLowercase{\textit{et al.}}: LIO-GVM: an Accurate, Tightly-Coupled Lidar-Inertial Odometry with Gaussian Voxel Map} 

\begin{abstract}
This letter presents a probabilistic voxel-based LiDAR Inertial Odometry framework for accurate and robust pose estimation. The framework addresses the correspondence mismatching issue by representing the LiDAR points as a set of Gaussian distributions and evaluating the divergence in variance for outlier rejection. Based on the fitted distributions, a new residual metric is proposed for the filter-based Lidar inertial odometry by incorporating both the distance and variance disparities, further enriching the comprehensiveness and accuracy of the residual metric. With the strategic design of the residual, we propose a simple yet effective voxel-solely mapping scheme, which only requires the maintenance of one centroid and one covariance matrix for each voxel. Experiments on different datasets demonstrate the robustness and high accuracy of our framework for various data inputs and environments. To the benefit of the robotics society, we open-source the code at \href{https://github.com/Ji1Xingyu/lio_gvm}{https://github.com/Ji1Xingyu/lio\_gvm}.
\end{abstract}


\section{Introduction}



\IEEEPARstart{T}{he} vital requirement for an autonomous robot is to estimate its ego-motion and build a map of the environment without prior knowledge, which is widely known as Simultaneous Localization and Mapping (SLAM). In particular, the LiDAR and IMU sensors are commonly employed together \cite{shan2020lio, lio_mapping, xu2021fast, xu2022fast, bai2022faster, qin2020lins} in SLAM problem to form a LiDAR Inertial Odometry (LIO) because of their complementary ability: LiDAR contributes accurate range measurements for the environment, while IMU provides precise and surrounding-insensitive short-term motion information \cite{ludwig2021investigation}. Because of the temporal and computational efficiency, filter-based LIO \cite{xu2021fast, xu2022fast, bai2022faster, qin2020lins} has been investigated in the recent few years. The performance of a LiDAR odometry is predominantly influenced by two components: point cloud registration and mapping scheme \cite{yuan2021survey}. In this letter, we aim to enhance the performance of a filter-based LIO system by addressing improvements to the shortcomings of the existing works in these two areas.

\begin{figure}[!t]
\centering
\includegraphics[width=3.0in]{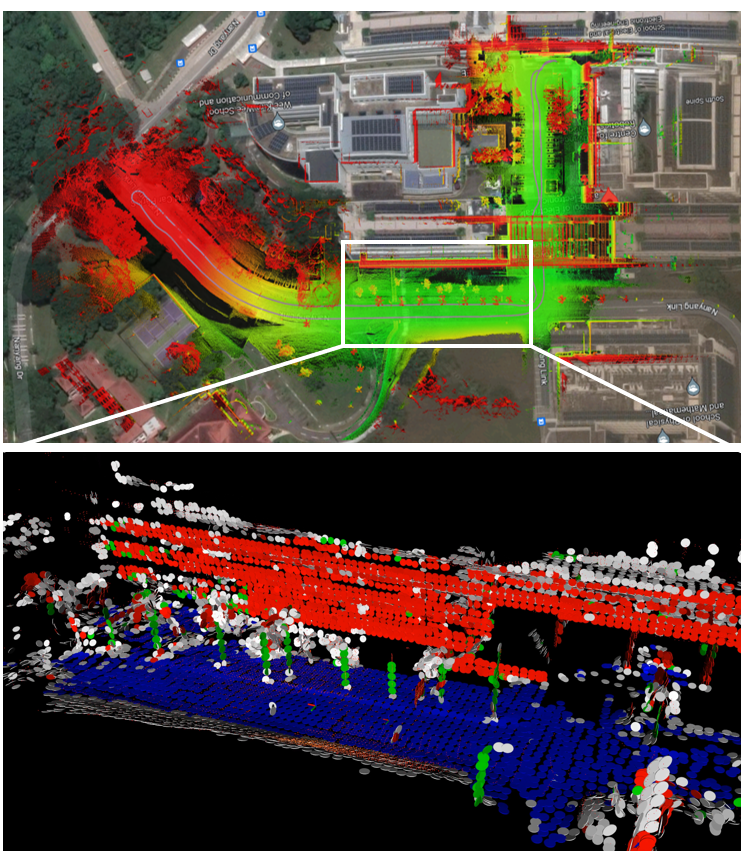}
\caption{The top image illustrates the global map built by LIO-GVM aligned with the Google map. The bottom image shows the runtime voxel map of LIO-GVM inside the rectangle. For better visualization, we model the distributions within each voxel as surfels. The blue, red, green, and white surfels denote the ground, wall, pole, and other types of voxels respectively.}
\label{fig:eee}
\vspace{-9pt}
\end{figure}

Point cloud registration refers to aligning the source LiDAR scan with a target map together, namely, to estimate the optimal transformation between their local coordinates. Early works extract the geometric feature points from the input scan and match them with the nearest point in the target map \cite{zhang2014loam, shan2018lego, f_loam}. A limitation of these methods is their deterministic target map representation, which cannot account for measurement uncertainties, resulting in inaccuracies in mapping and pose estimation. To address this issue, subsequent works model the target map in a probabilistic way, such as planes \cite{yuan2022efficient} or surfels \cite{suma, nguyen2023slict, surfel++}. The matched correspondences with large uncertainty will be depressed in the pose estimation phase. Nevertheless, they still employ the nearest correspondence matching. The key limitation of this matching scheme is its inability to reject false positive correspondences, which occurs when the source point is matched with the wrong target feature but identified as a valid pair. This issue can negatively affect the pose estimation accuracy. 

The mapping scheme acts as the interdependent component with point cloud registration, which maintains the target map for efficient environment representation. A common method is to maintain the target map with a k-d tree \cite{zhang2014loam, shan2018lego, f_loam} as it can provide strict K nearest neighbor (k-NN) search for the correspondence matching. The time complexity for the deletion, insertion, and query operations of the k-d tree is $\mathcal{O}(logN)$, which can impact real-time performance when dealing with a significantly large target map. To address this problem, some systems opt to govern the target map with hash-based voxels \cite{bai2022faster, yuan2022efficient, super_odom}, with $\mathcal{O}(1)$ the time complexity for element operation. However, due to the requirement of the front end, these voxel-based maps need to embed a supplemental data structure (k-d tree, Hilbert Curve, etc.) within voxels, which cannot reach the optimal operation speed.

To cope with the problem of mismatch in point cloud registration, we represent the points in both the source and target clouds as Gaussian distributions \cite{ndt, gicp, vgicp} and evaluate the matched correspondences by the disparity in variance. Moreover, based on the distribution assumption, we propose a new residual metric for filter-based LIO. The formula accesses both the distance and variance disparities, further enriching the comprehensiveness and accuracy of the residual metric. For the mapping side, instead of storing all the points, we propose an incremental voxel map without any additional structures, which only maintains a centroid and covariance matrix for each voxel and achieves $\mathcal{O}(1)$ time complexity. 

In conclusion, we implement the proposed point cloud registration and mapping scheme to a filter-based LIO system denoted as LIO-GVM. The contributions can be summarized as follows:
\begin{itemize}
    \item We propose a new correspondence matching method and residual metric for filter-based LIO. This matching scheme can reject false positive correspondences. Additionally, the metric measures not only distance but also variance disparity, enhancing the comprehensiveness and accuracy of the residual formulation. 
    \item We propose a simple yet efficient mapping scheme. The map only necessitates the maintenance of the centroid and covariance matrix for each voxel, which can achieve $\mathcal{O}(1)$ time complexity for element operation.
    \item We implement the two key techniques into a filter-based LIO system and evaluate its robustness and efficiency through various experiments (see Fig. \ref{fig:eee}). To the benefit of the robotics community, our implementation code is open-sourced at \href{https://github.com/Ji1Xingyu/lio_gvm}{https://github.com/Ji1Xingyu/lio\_gvm}.
\end{itemize}

\section{Related Work}

For point cloud registration, LOAM \cite{zhang2014loam} first proposes to extract planar and edge features from the source and target clouds for fast registration and achieves accurate results. LeGO-LOAM \cite{shan2018lego} performs Euclidean cluster first to extract the ground plane and restrict the correspondence match in the same cluster. Instead of extracting the geometric features from the source scan, some systems directly register the raw source cloud into the target map \cite{xu2022fast, xu2021fast}. SuMa \cite{suma} and SLICT \cite{nguyen2023slict} render points in the target map into surfels and consider the point-to-surfel distance as residual. Similarly, VoxelMap \cite{yuan2022efficient} proposes to maintain a probabilistic plane-based target map and register the point in the source scan to the map. These point-to-surfel (or plane) residuals consider the noise in the target map and thus can depress the influence of matched pairs with high uncertainty in the pose estimation phase. Nevertheless, they cannot handle the false positive correspondence problem. On the contrary, our proposed matching scheme can evaluate the similarity of the matched distributions and reject the false positive pairs.

The proposed residual metric is a weighted distribution-to-distribution (D2D) distance and is similar to the D2D residual in GICP \cite{gicp} and D2D-NDT \cite{d2d_ndt}. The key distinction lies in the proposed similarity-based weighting scheme, which can adjust the weight of the residual for matched correspondences based on their divergence in covariance. Subsequently, the proposed residual metric can provide a more comprehensive and accurate pose estimation. Similarly, LiTAMIN2 \cite{yokozuka2021litamin2} introduces symmetric KL-Divergence to the residual metric for divergence evaluation. However, their proposed formulation cannot be directly deployed in filter-based pose estimation.

For mapping schemes, a common method is to maintain the target map with a k-d tree \cite{zhang2014loam, f_loam}. The conventional k-d tree needs to rebuild the entire tree for the map update, which is inefficient and time-consuming. FAST-LIO2 \cite{xu2022fast} uses the ikd-Tree \cite{ikd_tree}, a dynamic k-d tree that can incrementally update the map. VoxelMap, Faster-LIO \cite{bai2022faster}, and SuperOdom \cite{super_odom} govern the target map with hash-based voxels. However, due to the requirement of the front end, these voxel-based mapping schemes need to embed a supplemental data structure within voxels to form a hierarchical structure for the maintenance of the raw points. Instead, the proposed mapping scheme only requires the maintenance of the centroid and covariance matrix for each voxel, resulting in a time complexity of $\mathcal{O}(1)$ for element operation. 

\section{Preliminary}

\begin{table}[!t]
\caption{Notations of this letter\label{tab:notation}}
\centering
\begin{tabular}{l@{\hspace{0.9em}}l}
    \toprule
Notation & Explanation \\ 
\midrule
$G, I, L$                                  &     The Global, IMU, and LiDAR frame            \\
$t_k$                                  &     The end time of the $k_{th}$ LiDAR scan \\
$\tau _i$                                  &     The sample time of $i_{th}$ IMU input, N.B. $\tau_i \in \left[ t_{k-1}, t_k \right]$          \\
${}^G\mathbf{T}_I$                         &     The transformation matrix of $I$ w.r.t. $G$            \\
${}^G\mathbf{R}_I,{}^G\mathbf{p}_I$        &     The rotation matrix and translation vector of $I$ w.r.t. $G$            \\
${}^{L_k}\mathbf{p}_j$                     &     The $j_{th}$ point coordinate with reference to $L$ at time $t_k$             \\
${}^G\mathbf{v}_i$                         &     The coordinate of the $i_{th}$ voxel with reference to $G$            \\
${}^G\boldsymbol{\mu}_i, {}^G\mathbf{C}_i$       &     The centroid vector and covariance matrix of voxel at ${}^G\mathbf{v}_i$             \\ 
$\left[ \mathbf{p} \right] _\times, \mathbf{p}^T$          &     The skew-symmetric matrix and transpose of vector $\mathbf{p}$ \\
\bottomrule
\end{tabular}
\vspace{-10pt}
\end{table}

\subsection{Notation}
The important notations used in this letter are listed in Tab. \ref{tab:notation}. Besides, we assume a static  calibrated extrinsic matrix ${}^{L}\mathbf{T}_{I} = ({}^{L}\mathbf{R}_{I}, {}^{L}\mathbf{p}_{I})$ from IMU frame to LiDAR frame.

\vspace{-5pt}

\subsection{System Description} \label{sec_sys_descr}
The proposed LIO-GVM is modeled as a discrete-time dynamical system sampled at the IMU rate. Its state $\mathbf{x}$ is defined on manifold $\mathcal{M}$ as:
\begin{equation} \label{eq_state_define}
\begin{split}
    \mathcal{M} &= SO(3)\times \mathbb{R}^{15}, \\ 
        \mathbf x &\stackrel{.}{=} 
        \setlength{\arraycolsep}{3.0pt}
    \begin{bmatrix}
    {}^G\mathbf{R}_I^T & {}^G\mathbf{p}_I^T & {}^G\mathbf{v}^T_I & \mathbf{b}^T_{\boldsymbol{\omega}} & \mathbf{b}^T_\mathbf{a} & {}^G\mathbf{g}^T
    \end{bmatrix}^T,
\end{split}
\end{equation}
where ${}^G\mathbf{v}_I$ and ${}^G\mathbf{g}$ are the velocity of IMU and gravity vector with reference to the global frame, $\mathbf{b}_{\boldsymbol{\omega}}$ and $\mathbf{b}_\mathbf{a}$ are the bias vectors of gyroscope and accelerometer of the IMU. The true state $\mathbf{x}$ of the system can be divided into two states: 
\begin{equation} \label{eq_truex}
    \begin{aligned}
        \mathbf{x}_k &= \hat{\mathbf{x}}_k \boxplus \delta \mathbf{x}_k, \\ 
        \delta \mathbf{x}_k &= \mathbf{x}_k \boxminus \hat{\mathbf{x}}_k =
        \setlength{\arraycolsep}{1.2pt}
        \begin{bmatrix}
            {}^G \delta \mathbf{r}_{I_k}^T & {}^G\delta\mathbf{p}_{I_k}^T & {}^G\delta\mathbf{v}^T_{I_k} & \delta\mathbf{b}^T_{\boldsymbol{\omega}_k} & \delta\mathbf{b}^T_{\mathbf{a}_k} & {}^G\delta\mathbf{g}_k^T
        \end{bmatrix} ^T, 
    \end{aligned}
\end{equation}
where $\delta \mathbf{x}_k \in \mathbb{R}^{18}$ is the error state, $\hat{\mathbf{x}}_k \in \mathcal{M}$ is the nominal state, $\boxminus$ and $\boxplus$ are encapsulated operators that represent a bijective mapping between the manifold $\mathcal{M}$ and its local tangent space $\mathbb{R}^{18}$ \cite{hertzberg2013integrating}; ${}^G \delta \mathbf{r}_{I_k} = {}^G\mathbf{R}_{I_k} \boxminus {}^G\hat{\mathbf{R}}_{I} = Log({}^G\mathbf{R}_{I_k}{}^G\hat{\mathbf{R}}_{I_k}^{-1})$ is the minimal representation of the error rotation. In this way, we can use the iterative error state Kalman filter (IESKF) to estimate the error state $\delta \mathbf{x}_k$ on Euclidean space and then project it back to the manifold using \eqref{eq_truex} for the maintenance of the true state $\mathbf{x}_k$. 

\vspace{-7pt}

\section{Methodology} \label{sec_method}

\begin{figure*}[!t]
\centering
\includegraphics[width=5.5in]{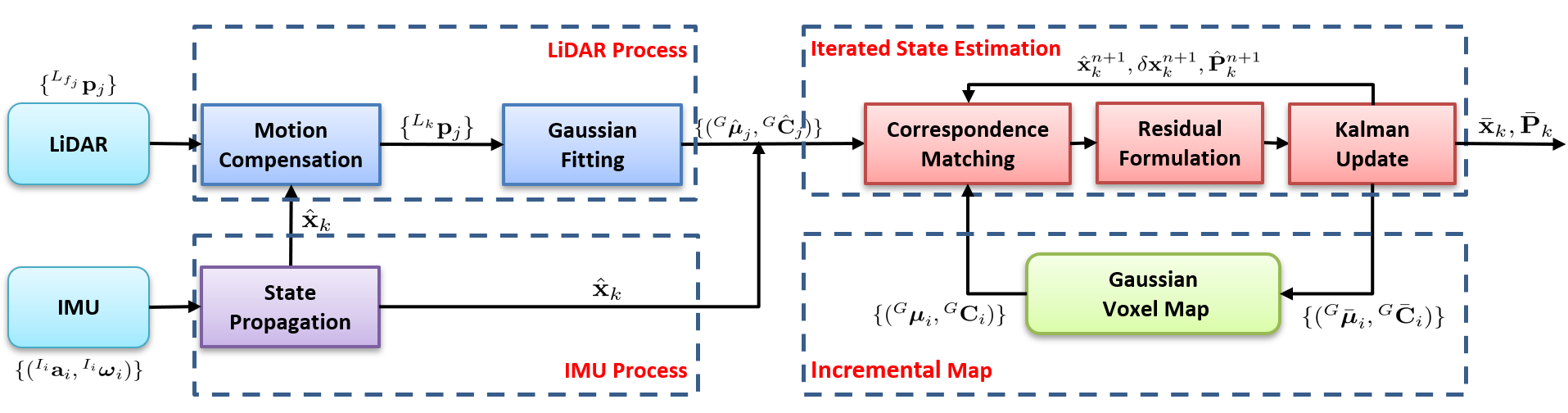}
\caption{System Workflow.}
\label{fig_workflow}
\vspace{-7pt}
\end{figure*}
Given the estimate of the last scan: $\Bar{\mathbf{x}}_{k-1}$ and $\Bar{\mathbf{P}}_{k-1}$, we will detail the system workflow in the following subsections.

\vspace{-7pt}

\subsection{IMU Process}
Recalling the state defined in \eqref{eq_state_define}, we have the following discrete model based on continuous kinematics equations \cite{hertzberg2013integrating}:
\begin{equation} \label{eq_true_pred}
            {\mathbf{x}}_{i+1} = {\mathbf{x}}_i \boxplus \mathbf{f}({\mathbf{x}}_i, \mathbf{u}_i, \mathbf{w}_i),
\end{equation}
where $\mathbf{u}_i$, $\mathbf{w}_i$, and $\mathbf{f}$ are the system input, system noise, and kinematics function respectively. Please refer to Appendix \ref{sec_apd_a} for the detailed formulations. Firstly, the mean of the nominal state $\hat{\mathbf{x}}_i$ will be predicted without considering the noise $\mathbf{w}_i$: 
\begin{equation} \label{eq_nominal_pred}
           \hat{\mathbf{x}}_{i+1} = \hat{\mathbf{x}}_{i} \boxplus \mathbf{f}(\hat{\mathbf{x}}_{i}, \mathbf{u}_{i}, \mathbf{0}); \; \hat{\mathbf{x}}_{0} = \Bar{\mathbf{x}}_{k-1}.
\end{equation}
Consequently, this will result in the accumulation of inaccuracies. To cope with the problem, $\mathbf{w}_i$ is taken into account in the prediction of the error state $\delta \mathbf{x}_i$:
\begin{equation}
    \delta \mathbf{x}_{i+1} \approx \mathbf{F}_{\delta \mathbf{x}} \delta \mathbf{x}_i + \mathbf{F}_{\mathbf{w}} \mathbf{w}_i; \delta \mathbf{x}_{0} = \delta \Bar{\mathbf{x}}_{k-1} = \mathbf{0}, \label{eq_ex_pred} \\
\end{equation}
where $\mathbf{F}_{\delta \mathbf{x}}$ and $\mathbf{F}_{\mathbf{w}}$ are the Jacobian matrices of $\mathbf{f}()$ w.r.t. $\delta \mathbf{x}_i$ and $\mathbf{w}_i$ \cite{xu2022fast}. In the implementation, the system noise $\mathbf{w}_i$ is considered as Gaussian white noise following the normal distribution $\mathcal{N}(\mathbf{0}, \mathbf{W})$, where $\mathbf{W}$ is the covariance matrix. The error state $\delta \mathbf{x}_i$ is an independent random vector with multivariate normal distribution: $\delta \mathbf{x}_i\sim \mathcal{N}\left( \mathbf{0}, \mathbf{P}_i \right)$. As a consequence, the predicted covariance matrix $\hat{\mathbf{P}}_{i+1}$ of $\delta \mathbf{x}_{i+1}$ can be calculated as the linear transformation:
\begin{equation}
        \hat{\mathbf{P}}_{i+1} = \mathbf{F}_{\delta \mathbf{x}} \hat{\mathbf{P}}_{i} \mathbf{F}_{\delta \mathbf{x}}^T + \mathbf{F}_{\mathbf{w}} \mathbf{W} \mathbf{F}_{\mathbf{w}}^T; \; \hat{\mathbf{P}}_0 = \Bar{\mathbf{P}}_{k-1}. \label{eq_p_pred}
\end{equation}
The IMU process module utilizes the state estimate from the previous scan as the initial condition and recursively predicts $\hat{\mathbf{x}}_{i+1}$, $\delta \mathbf{x}_{i+1}$ and $\hat{\mathbf{P}}_{i+1}$ till the end time of the current LiDAR scan $t_k$. 
\subsection{LiDAR Process} \label{sec_lidar}
All the LiDAR points ${}^{L_j}\mathbf{p}_j$ with timestamps $\gamma_j \in \left[ t_{k-1}, t_k \right]$ are cached to form a point cloud $\mathcal{P}_k$. For most commercial LiDAR, the points are sampled under different timestamps, which will introduce motion distortion to the cloud $\mathcal{P}_k$. We seek to compensate for the distortion by projecting all the points in $\mathcal{P}_k$ to $L_k$, the LiDAR frame at $t_k$. To this end, we adopt the backward propagation proposed in \cite{xu2021fast} to obtain the relative motion between IMU frames sampled at $\gamma_j$ and $t_k$: ${}^{I_k}\mathbf{T}_{I_j}$. The point ${}^{L_j}\mathbf{p}_j$ can thus be projected to $L_k$ as: 
\begin{equation} 
    {}^{L_k}\mathbf{p}_j = {}^IT_L^{-1} {}^{I_k}\mathbf{T}_{I_j} {}^IT_L {}^{L_j}\mathbf{p}_j.
\end{equation}
Moreover, a representative descriptor for the point is required for robust correspondence matching. Thus, we represent the deskewd points as a collection of Gaussian distributions. For each point ${}^{L_k}\mathbf{p}_j$, we collect its nearest $y$ neighbors and fit them to a Gaussian distribution $\mathcal{N}({}^{L_k}\boldsymbol{\mu}_j, {}^{L_k}\mathbf{C}_j)$ \cite{ndt}. Eventually, the LiDAR process module outputs a collection of the fitted Gaussian distributions for all the points: $\{ {}^{L_k}\boldsymbol{\mu}_j, {}^{L_k}\mathbf{C}_j \}$.

\vspace{-12pt}

\subsection{Iterated State Estimation}
\begin{figure}[!t]
\centering
\includegraphics[width=2.8in]{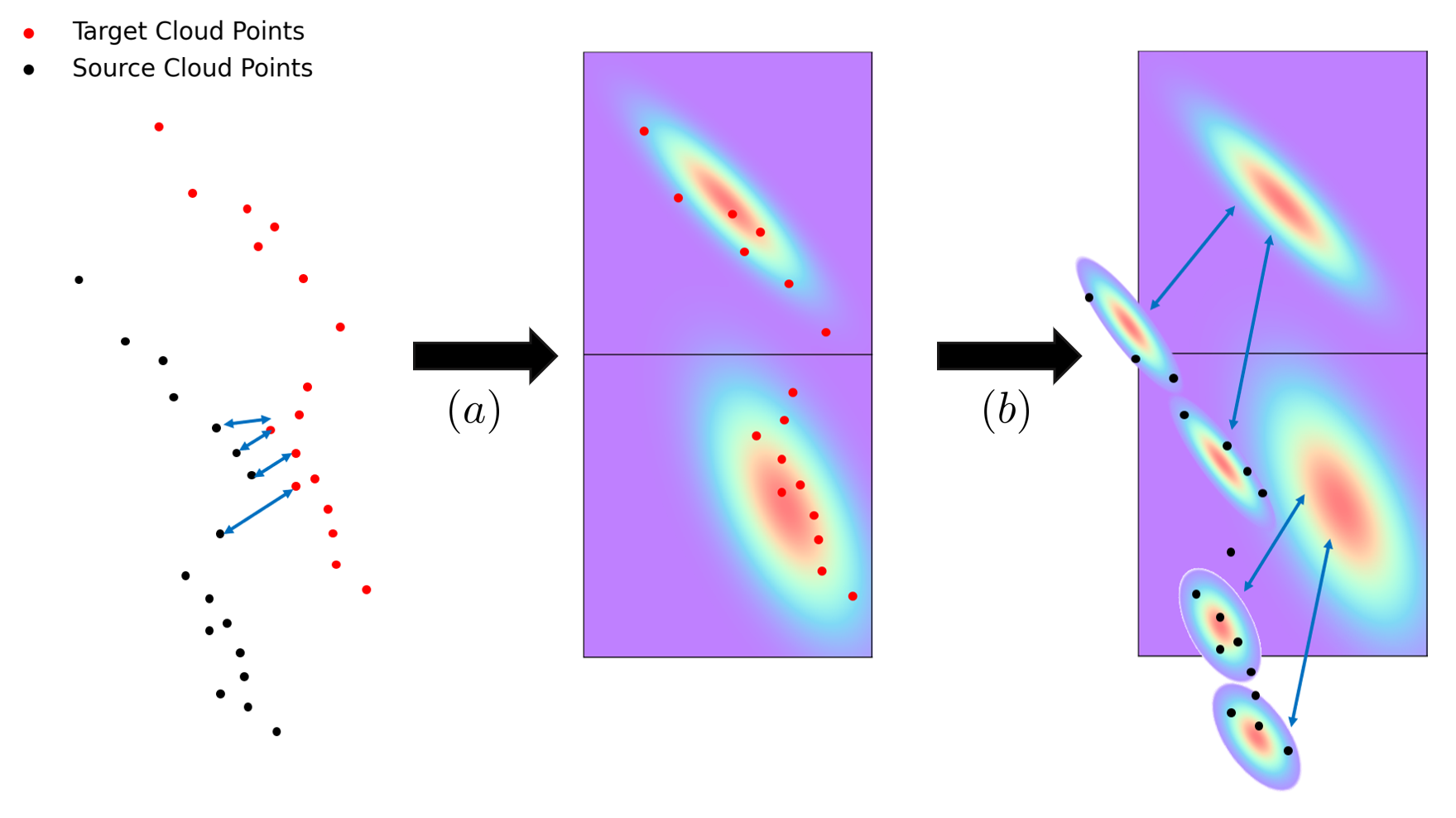}
\caption{Directly registering the source points to the nearest target points would generate mismatches (shown in the left image). Instead, we: (a). Voxelize the target cloud and fit the points within each voxel with a Gaussian distribution. (b). Fit each source point and its nearest $y$ neighbors with a Gaussian distribution and match it with the near \& similar target distributions.}
\label{fig:match}
\vspace{-10pt}
\end{figure}

Upon completion of the IMU and LiDAR processes, the following steps for the IESKF-based LIO system involve correspondence matching, residual formulation, and Kalman update. However, existing filter-based LIO systems \cite{xu2021fast, qin2020lins, bai2022faster} typically adopt the nearest correspondence matching strategy and treat the ensuing point-to-feature distance as the residual. This strategy involves projecting ${}^{L_k}\mathbf{p}_j$ to the global frame as:
\begin{equation} \label{eq_pL2G}
        {}^{G}\hat{\mathbf{p}}_j = {}^G\hat{\mathbf{T}}_{I_{k}} {}^IT_L {}^{L_k}\mathbf{p}_j
        = {}^G\Bar{\mathbf{T}}_{I_{k-1}} {}^{I_{k-1}}\hat{\mathbf{T}}_{I_k} {}^IT_L {}^{L_k}\mathbf{p}_j.
\end{equation}
For long-term odometry, the inevitable accumulation of localization drifts introduces errors to ${}^G\Bar{\mathbf{T}}_{I_{k-1}}$. Simultaneously, the presence of spike noise, a common issue in low-cost IMUs, leads to inaccuracies in ${}^{I_{k-1}}\hat{\mathbf{T}}_{I_k}$. These factors prevent ${}^{L_k}\mathbf{p}_j$ from being projected to its exact corresponding feature, generating mismatches (see Fig. \ref{fig:match}). Besides, matching the nearest feature in the map could be computationally intensive if the map size grows too large in long-term odometry. To cope with these issues, we propose a new residual formulation for IESKF, which can mitigate incorrectly matched correspondences while maintaining computational efficiency (see Fig. \ref{fig:match}). The details are described in the following subsections. 

\subsubsection{Correspondence Matching} \label{sec_match}

\begin{figure}[!t]
\centering
\includegraphics[width=2.7in]{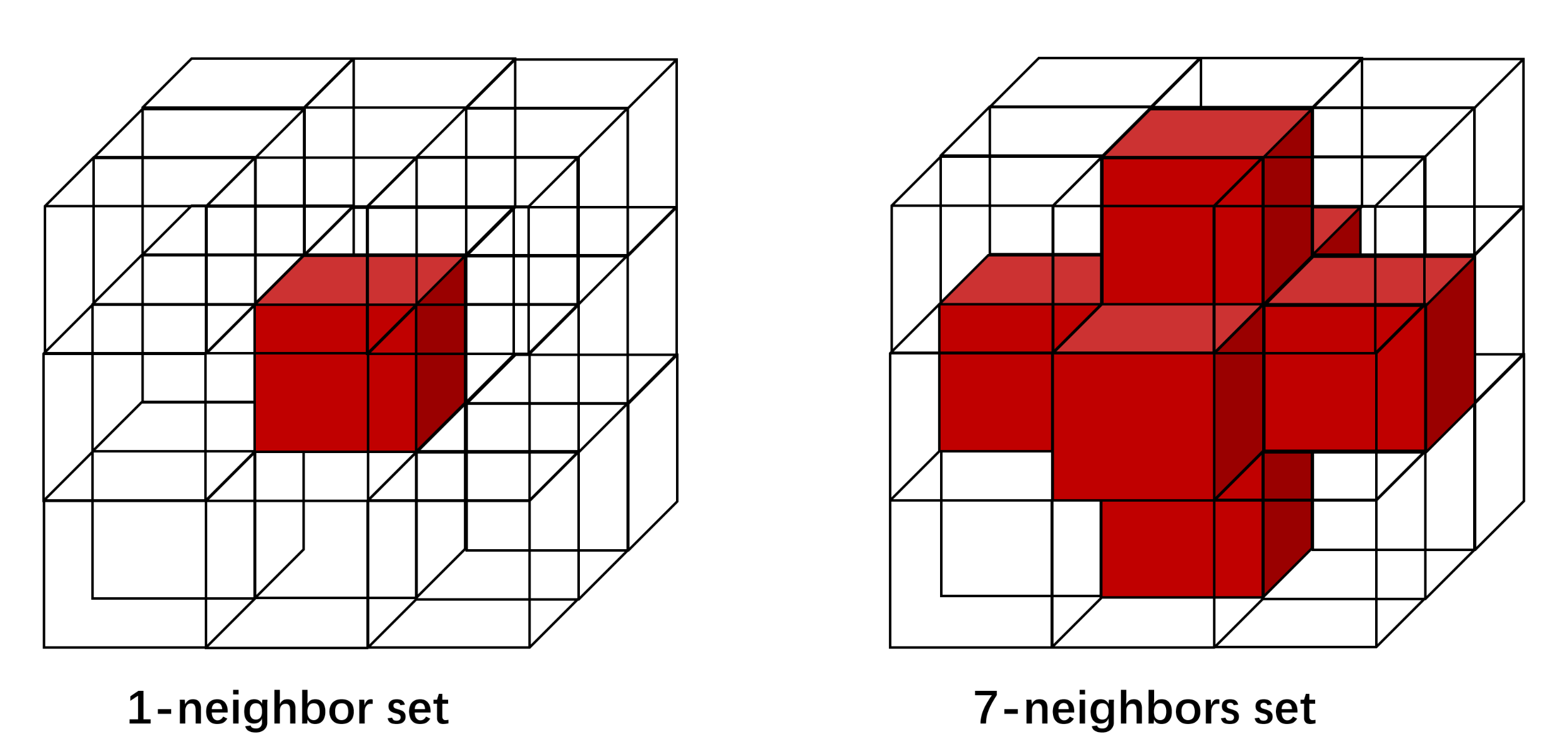}
\caption{The different neighbor set $\mathcal{V}_j$ of a given voxel ${}^G\mathbf{v}_j^n$.}
\label{fig_voxel}
\vspace{-12pt}
\end{figure}

Suppose the iterated state estimation module is under the $n_{th}$ iteration and the corresponding state estimation is $\hat{\mathbf{x}}^n_k, \delta \mathbf{x}^n_k$. We first project the fitted distributions into the global frame as: 
\begin{equation}
\begin{aligned}
    {}^{G}\hat{\boldsymbol{\mu}}_j^n &= {}^G \hat{\mathbf{T}}_{L_k}^n {}^{L_k}\boldsymbol{\mu}_j, \\
    {}^{G}\hat{\mathbf{C}}_j^n &= {}^G\hat{\mathbf{R}}_{L_k}^n {}^{L_k}\mathbf{C}_j {{}^G\hat{\mathbf{R}}_{L_k}^n}^T.
\end{aligned}
\end{equation}
For each projected distribution $\left( {}^{G}\hat{\boldsymbol{\mu}}_j^n, {}^{G}\hat{\mathbf{C}}_j^n \right)$, we aim to find its near and similar correspondence in the global voxel map. Each voxel contains one Gaussian distribution $\left( {}^G\boldsymbol{\mu}_i, {}^G\mathbf{C}_i\right)$ (details of the global map is presented in Section \ref{sec_map}). Hellinger distance is a statistical measure that quantifies the divergence between two probability distributions. Its value is dominated by two terms: a similarity term and a distance term. The similarity term captures the resemblance between two probability distributions, while the distance term quantifies the separation. For the correspondence evaluation, we extract the similarity term from the squared Hellinger distance \cite{gibbs2002choosing} and denote it as the similarity metric $s_{j,i}$: 
\begin{equation}
    s_{j,i} = \sqrt{\frac{\left(\det{\mathbf{C}_j} \det{\mathbf{C}_i}\right)^{1/2}}{\det{(\mathbf{C}_j/2 + \mathbf{C}_i/2)}}}.  
\end{equation} 
The larger the $s_{j,i}$ is, the more similar the two distributions are. The correspondence matching procedure can now be conceptualized as follows: We first find the voxel coordinate ${}^G\mathbf{v}_j^n$ in the global map in which ${}^{G}\hat{\boldsymbol{\mu}}_j^n$ occupies. Then, we define a neighbor set of the occupied voxel as $\mathcal{V}_j$ (the selection for $\mathcal{V}_j$ is illustrated in Fig. \ref{fig_voxel}). After that, we calculate the similarity $s_{j,i}$ between the source distribution and the target distribution in ${}^G\mathbf{v}_i$ (${}^G\mathbf{v}_i \in \boldsymbol{\mathcal{V}}_j$) and discard the pairs with $s_{j,i}$ smaller than a threshold $s_t$. Empirically, the selection for $s_t$ is influenced by the noise level of the LiDAR and IMU: the larger the noise is, the smaller the $s_t$ should be selected. Generally, $s_t$ should not be selected in a strict manner, as the noise inherent in the LiDAR sampling and deskewing procedures could potentially have an adverse impact on similarity.

\subsubsection{Residual Formulation}
Before presenting the new residual metric, it is necessary to explain the observation model and state estimation scheme for IESKF. This will provide the rationale behind the proposed residual metric. The observation function of the system is: 
\begin{equation}
\begin{aligned}
    \mathbf{z}_k^n &=  \mathbf{h}\left( \hat{\mathbf{x}}_k^n \boxplus \delta \mathbf{x}_k^n \right) + \mathbf{v}_k, 
\end{aligned}
\end{equation}
where $\mathbf{v}_k$ is the measurement noise satisfying $\mathbf{v}_k \sim n\left( 0, \mathbf{V}_k \right)$, $\mathbf{h}\left( \right)$ is the observation function that maps the state space to the observation space. Since the error state is quite small, we can linearize the observation at $\hat{\mathbf{x}}_k^n$:
\begin{equation}
        \mathbf{z}_k^n \approx \mathbf{h}\left( \hat{\mathbf{x}}_k^n \right) + \mathbf{H}^n\delta \mathbf{x}_k^n + \mathbf{v}_k \label{eq_pos},
\end{equation}
where $\mathbf{H}^n$ is the Jacobian matrix of $\mathbf{h}\left( \hat{\mathbf{x}}_k^n \boxplus \delta \mathbf{x}_k^n \right)$ w.r.t. $\delta \mathbf{x}_k^n$. As $\delta \mathbf{x}_k$ and $\delta \mathbf{x}_k^n$ lie in different tangent space of the manifold $\mathcal{M}$ (the tangent space of $\hat{\mathbf{x}}_k$ and $\hat{\mathbf{x}}_k^n$ respectively), we need to project $\delta \mathbf{x}_k$ to $\delta \mathbf{x}_k^n$ in the tangent space as: 
\begin{equation} \label{eq_pri}
    \delta \mathbf{x}_k = \left( \hat{\mathbf{x}}_k^n \boxplus \delta \mathbf{x}_k^n \right) \boxminus \hat{\mathbf{x}}_k =  \hat{\mathbf{x}}_k^n \boxminus \hat{\mathbf{x}}_k + \left( \mathbf{J}^n\right)^{-1}\delta \mathbf{x}_k^n, \\
\end{equation}
where $\mathbf{J}^n$ is the inverse of the Jacobian matrix of $\left( \hat{\mathbf{x}}_k^n \boxplus \delta \mathbf{x}_k^n \right) \boxminus \hat{\mathbf{x}}_k$ w.r.t. $\mathbf{x}_k^n$. Readers may refer to Appendix \ref{sec_apd_a} for the detailed formulation of $\mathbf{J}^n$. Combining \eqref{eq_pos} and \eqref{eq_pri}, following the Bayesian rule, we have the maximum a posterior (MAP) problem:
\begin{equation} \label{eq_map}
        \delta \mathbf{x}_k^n = \arg\min_{\delta \mathbf{x}_k^n} \left( \| \delta \mathbf{x}_k \|^2_{\hat{\mathbf{P}}_k^{-1}} + \| \mathbf{h}\left( \hat{\mathbf{x}}_k^n \right) + \mathbf{H}^n\delta \mathbf{x}_k^n \|^2_{\mathbf{V}_k^{-1}} \right).
\end{equation}

Given that the matched correspondences are Gaussian distribution pairs, we aim to design a residual metric such that the second term in \eqref{eq_map} can represent both the distance and the variance disparity. Let's consider the weighted Mahalanobis distance presented as follows:
\begin{equation} \label{eq_weighted_md}
    \sum_{j=0}^m (s_{j,i}^n)^2 \left( {}^{G}\hat{\boldsymbol{\mu}}_j^n - {}^G\boldsymbol{\mu}_i \right)^T (\hat{\mathbf{C}}_j^n + \mathbf{C}_i + \alpha \mathbf{I})^{-1} \left( {}^{G}\hat{\boldsymbol{\mu}}_j^n - {}^G\boldsymbol{\mu}_i \right), 
\end{equation}
where $m$ is the number of matched correspondences, $\alpha \mathbf{I}$ ($\alpha > 0$) is a constant diagonal matrix to avoid the singularity issue. The Mahalanobis distance is weighted by the similarity of the matched correspondences, ensuring that the pairs with higher similarity and closer distance contribute more to the estimation process, which aligns with our objective. Adopting this formula, the resulting residual metric for $\left( {}^{G}\hat{\boldsymbol{\mu}}_j^n, {}^{G}\hat{\mathbf{C}}_j^n \right)$ would be: 
\begin{equation}
    \mathbf{z}_{k,j}^n = s_{j,i}^n\mathbf{D}_j^n \left( {}^{G}\hat{\boldsymbol{\mu}}_j^n - {}^G\boldsymbol{\mu}_i \right). 
\end{equation}
Appendix \ref{sec_res_proof} exhibits the detailed derivation and definition. 
\subsubsection{State Update}
The Jacobian matrix $\mathbf{H}_{k,j}^n$ of $\mathbf{z}_{k,j}^n$ w.r.t. $\delta \mathbf{x}_k^n$ is:
\begin{equation}
    \mathbf{H}_{k,j}^n = \begin{bmatrix}
        -{s_j} {\mathbf{D}_j} \left[  {}^I\mathbf{q}_j \right]_\times & {s_j} {\mathbf{D}_j} & \mathbf{0}_{3\times 15}
    \end{bmatrix}.
\end{equation}
Please refer to Appendix \ref{sec_apd_a} for the derivation and definition details. With the Jacobian matrix, the optimal solution of the MAP problem \eqref{eq_map} can be solved by the Kalman update \cite{he2021kalman} described in Appendix \ref{sec_apd_a}. 
\vspace{-10pt}

\subsection{Incremental Global Map} \label{sec_map}
The global map is composed of voxels stored as a hash table: $\mathbb{Z}^3 \rightarrow \mathbb{R}^3 \times \mathbb{R}^{3\times3}$, which builds a map from the voxel coordinate ${}^G\mathbf{v}_i$ to its centroid ${}^G\boldsymbol{\mu}_i$ and the covariance matrix ${}^G\mathbf{C}_i$. This structure demonstrates significant efficiency in querying, achieving an optimal $\mathcal{O}(1)$ time complexity for correspondence matching. This structure is similar to that of VGICP \cite{vgicp}. In contrast to VGICP, which only supports pair-wise registration and requires repeated voxelization of the source scan, our approach offers a complete solution for insert, update, and delete operations within a global target map. We also address the variable LiDAR sampling rate problem (e.g. the ground voxels are undersampled in the distance and degrade from the "plane" to "line" feature) by introducing an update scheme to ensure the continuous and accurate maintenance of the map.

For the initialization, the first scan is regarded as the global frame $G = L_0$. Each point in the scan is subsequently fitted to a Gaussian distribution as the procedure in \ref{sec_lidar} to obtain $\left( {}^{G}\hat{\mathbf{p}}_j, \mathbf{C}_j \right)$. Consequently, the scan is voxelized with voxel size $r$. Suppose we have the voxel with coordinate ${}^G\mathbf{v}_i$ encompassing a set of points  $\{{}^G\mathbf{p}_m\}$, the centroid and covariance matrix ${}^G\boldsymbol{\mu}_i, {}^G\mathbf{C}_i$ will be calculated as the average of the distributions within. Once the current estimate $\Bar{\mathbf{x}}_k$ is acquired, the current scan contributes to the global map. Since the projected Gaussian distributions have been obtained during the state estimation phase, we proceed to voxelize the current scan following \eqref{eq_voxelize}, resulting in a temporary voxel map ${}^G\mathbf{v}_j \rightarrow (\Bar{\boldsymbol{\mu}}_j, \Bar{\mathbf{C}}_j)$. For each voxel present in this temporary voxel map, we query its coordinate ${}^G\mathbf{v}_j$ in the global map. If the global map doesn't contain the voxel, we directly insert the voxel into the global map. This operation is time-efficient since the insert operation for a hash table has a time complexity of $\mathcal{O}(1)$. Otherwise, the global map will be updated as:
\begin{equation} \label{eq_voxelize}
    \begin{aligned}
        {}^G\boldsymbol{\mu}_j &= \frac{M{}^G\boldsymbol{\mu}_j + N{}^G\Bar{\boldsymbol{\mu}}_j}{M+N}, \\
        {}^G\mathbf{C}_j &= \frac{M{}^G\mathbf{C}_j + N{}^G\Bar{\mathbf{C}}_j}{M+N}, \\
    \end{aligned}
\end{equation}


where $M$ is the recorded point number of voxel ${}^G\mathbf{v}_j$ in the global map, $N$ is the corresponding one in the temporary voxel map. In this way, the global map only necessitates the maintenance of the centroid and covariance matrix for each voxel, which is considerably efficient in terms of memory usage and time efficiency compared to maintaining a tree structure storing all points \cite{xu2022fast} or embedding additional structures into the voxels \cite{bai2022faster, super_odom}. Specifically, $M$ is updated using the formula $M = max(M, N)$. This is motivated by the goal of maintaining a continuously updated voxel scheme. This approach ensures that $M$ is constrained by the maximum number of points that can be sampled by LiDAR within a voxel. Otherwise, recalling \eqref{eq_voxelize}, an unbounded $M$ would lead to $M \gg N$ after a few scans, thus preventing the update of voxel parameters. Consequently, this enables an incremental update of the global map without the necessity to delete voxels from the map. Nevertheless, the mapping scheme also provides the option to remove voxels out of the effective range of the LiDAR for operation in exceptionally large environments.

\section{Experiment}
In this section, the proposed system will be evaluated in terms of accuracy, temporal efficiency, and storage efficiency. All the experiments are conducted on a laptop with a 2.3 GHz AMD Ryzen 7 6800H CPU and 16 GB RAM. We chose 9 sequences from two representative datasets for the evaluation. The first 5 sequences, denoted as \textbf{nc\_*},  are selected from Newer College dataset \cite{newer_dataset}, with an OS1-64 LiDAR at a scan rate of 10Hz and a built-in 6-axis IMU sampled at 100Hz. The rest sequences are selected from the MCD VIRAL dataset \cite{mcdviral2023}, collected by an OS1-128 LiDAR sampled at 10Hz, a Livox Mid-70 LiDAR, and a VectorNav VN100 9-axis IMU sampled at 800 Hz. In order to evaluate the influence of IMU on the odometry accuracy, we perform experiments on different IMUs for the same sequences from MCD VIRAL dataset: \textbf{ntu\_*} denoting OS1-128 LiDAR with VN100 IMU, \textbf{ntuo\_*} denoting OS1-128 LiDAR with its built-in IMU. The ground truth poses for both datasets are generated by registering each LiDAR scan to a highly accurate prior map using an ICP method. 

\begin{table*}[!t]
\caption{evaluation and comparison of odometry accuracy on all the sequences}
\centering
\label{table:odom}
\resizebox{0.9\textwidth}{!}{
\begin{threeparttable}
\begin{tabularx}{\textwidth}{>{\raggedright\scriptsize} p{1.8cm}X*{14}{>{\centering\arraybackslash}X}}
\toprule
         & ntu\_02          & ntu\_04          & ntu\_10          & ntu\_13               & ntuo\_02         & ntuo\_04         & ntuo\_10              & ntuo\_13              & nc\_01     & nc\_02     & nc\_05       & nc\_06        & nc\_07 \\ 
\midrule                                                                                                                                                    
DLO      & 6.46/{0.43}                       & 3.65/\underline{0.11}           & 7.65/0.51                       & {3.98}/{0.20}            & 6.46/\underline{0.43}              & 3.75/\textbf{0.11}  & 7.65/0.51            & \underline{3.98}/\underline{0.20}            & 4.20/0.51       & 4.27/\underline{0.37}       & 3.44/0.44         & 1.31/0.62          & 1.73/0.18 \\
VoxelMap  & \underline{1.83}/\textbf{0.11}                                & 2.98/0.18                       & 8.48/0.57                       & $\times$            & \textbf{1.83}/\textbf{0.11}                & 2.98/0.18                & 8.48/0.57                     & $\times$                     & $\times$          & $\times$          & $\times$            & $\times$             & $\times$ \\
LIO-SAM  & 7.80/0.52                                & 3.71/0.23                       & 9.32/0.63                       & 3.05/0.16            & -$^a$                & -                & -                     & -                     & -          & -          & -            & -             & - \\
FAST-LIO2 & 2.09/\underline{0.24}       & \textbf{2.11}/0.13              & \textbf{2.34}/{0.15}  & {2.89}/\underline{0.11} & {5.64}/0.80       & \underline{3.51}/0.23           & \underline{7.32}/{0.41}  & 6.79/0.50            & \underline{3.23}/{0.37}       & \underline{4.15}/0.43       & \textbf{3.13}/\textbf{0.26}         & \underline{0.94}/\underline{0.37}       & \underline{1.16}/\textbf{0.15} \\
LIO-GVM(w/o s)   & 2.37/0.40       & 2.35/\underline{0.11}  & 3.21/\underline{0.14}  & \underline{2.79}/{0.14}  & {5.51}/{0.68}  & {4.73}/{0.39}  & {6.59}/\underline{0.39}  & 4.58/{0.26}  & {3.36}/\underline{0.31}  & {4.76}/{0.44}  & {3.46}/{0.47}  & {1.28}/{0.43}     & {1.38}/{0.20} \\
LIO-GVM(plane)   & 8.24/0.68      & 6.91/0.41  & 8.29/0.74  & 5.77/0.43  & 9.34/0.90  & 7.95/0.54  & 10.2/0.81  & 8.89/0.73  & 6.29/0.49  & 6.30/0.67  & 4.54/0.72  & 2.48/0.59     & 3.45/0.30 \\
LIO-GVM   & \textbf{1.74}/0.46                         & \underline{2.26}/\textbf{0.05}  & \underline{3.16}/\textbf{0.11}  & \textbf{2.74}/\textbf{0.09}  & \underline{3.23}/\underline{0.40}  & \textbf{2.69}/\underline{0.14}  & \textbf{4.25}/\textbf{0.27}  & \textbf{3.60}/\textbf{0.18}  & \textbf{2.47}/\textbf{0.26}  & \textbf{3.20}/\textbf{0.35}  & \underline{3.27}/\underline{0.33}  & \textbf{0.89}/\textbf{0.33}     & \textbf{1.10}/\underline{0.17} \\
\bottomrule
\end{tabularx}
\begin{tablenotes}
    \footnotesize
        \item Errors are denoted as \textbf{ATE[\%]/ARE[deg/10m]} (smaller is better). We only align the position and orientation at the origin for a comprehensive evaluation. 
        \item \textbf{Bold} value stands for the best; \underline{underline} is the second best;  '-' denotes that the algorithm is not applicable; '$\times$' indicates that the algorithm diverges.
        \item '-$^a$': LIO-SAM is not applicable for \textbf{ntuo\_*} an \textbf{nc\_*} sequences due to the lack of a 9-axis IMU.
\end{tablenotes}
\end{threeparttable}
}
\vspace{-7pt}
\end{table*}

\subsection{Evaluation of Odometry Accuracy}
We evaluate the accuracy by comparing the KITTI metric \cite{kitti_metric}: the average translation error (ATE), and the average rotation error (ARE). We compare LIO-GVM with three state-of-the-art LiDAR (-inertial) odometry algorithms: DLO \cite{direct_lidar_odom}, VoxelMap \cite{yuan2022efficient}, LIO-SAM \cite{shan2020lio}, and FAST-LIO2 \cite{xu2022fast}. DLO is a GICP-based LO, VoxelMap is a probabilistic voxel-based LO while LIO-SAM and FAST-LIO2 are LIO systems. Parameters for all methods are kept constant across sequences for consistency. For LIO-GVM, the voxel size, $s_t$, and $\alpha$ are empirically set as 1.0 $m$, 0.70 and $10^{-6}$, respectively. Additionally, the neighbor searching scheme is set as 7-neighbors searching.

\begin{figure}[!t]
\centering
\includegraphics[width=3.0in]{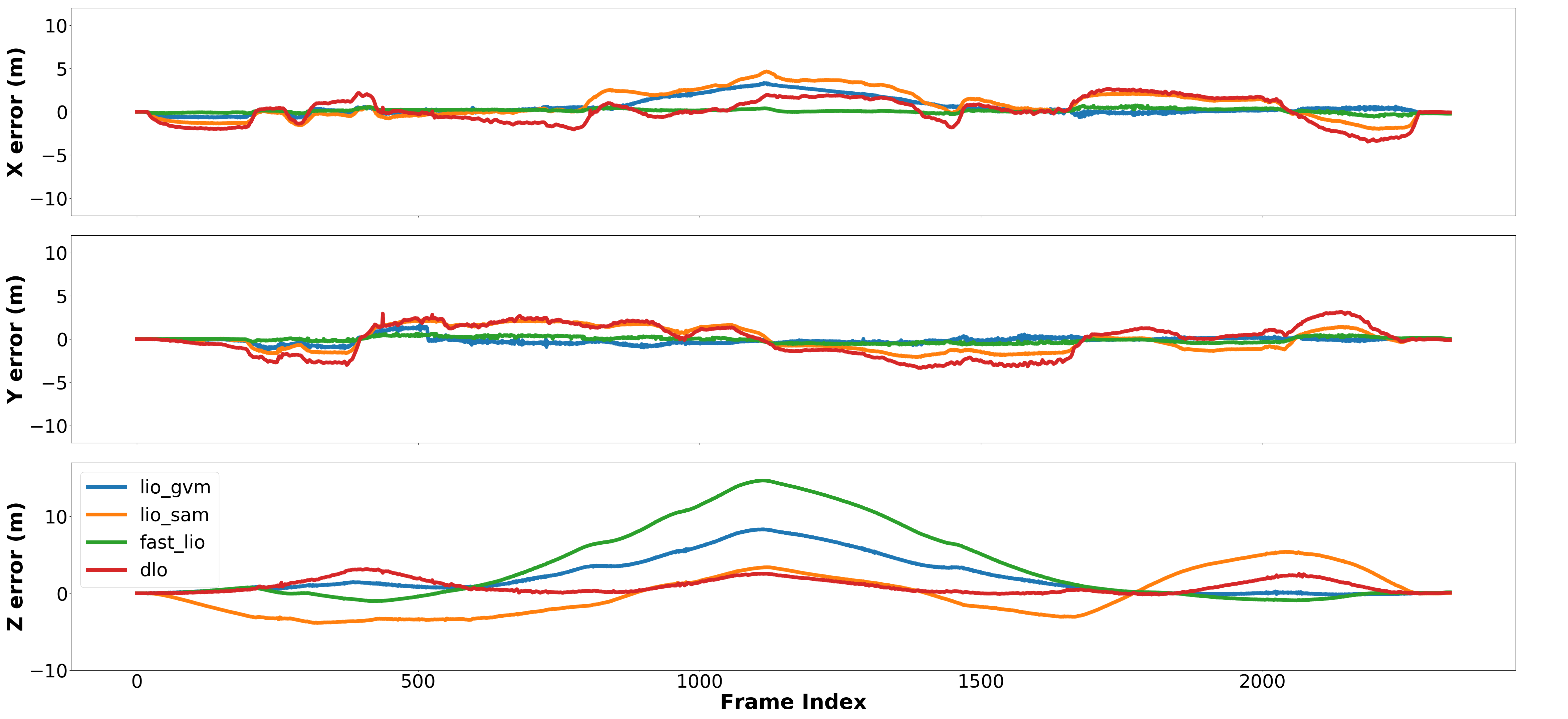}
\caption{The position error of different methods on sequence {ntu\_13}. We find that the position error is most significant along the z axis, due to the severe elevation changes.}
\label{fig:xyz}
\vspace{-10pt}
\end{figure}

\begin{figure}[!t]
\centering
\includegraphics[width=3.0in]{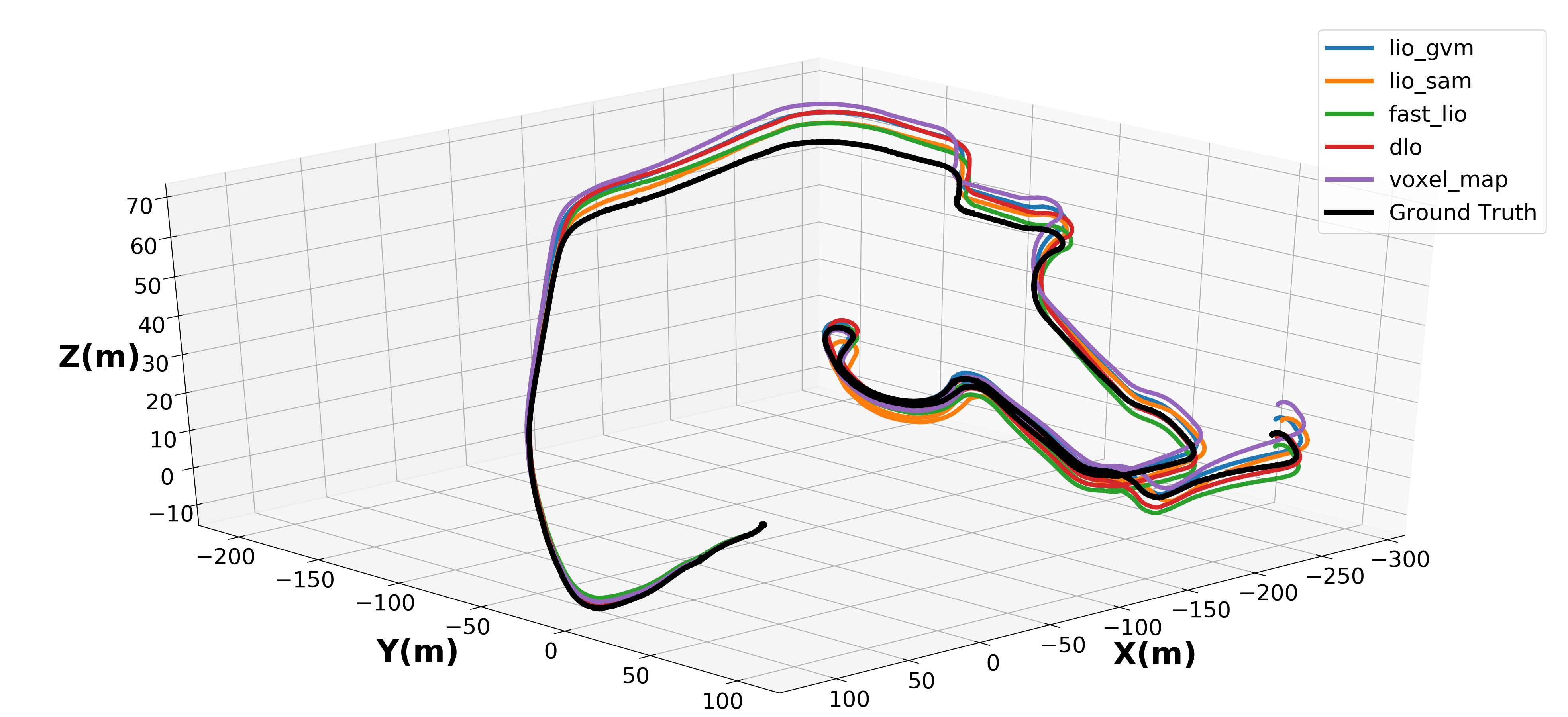}
\caption{The trajectories reference to the ground on sequence {ntu\_04}. }
\label{fig:traj}
\vspace{-10pt}
\end{figure}

Two variations of the proposed system denoted as LIO-GVM(w/o s) and LIO-GVM(plane) are tested for ablation study. LIO-GVM(w/o s) is implemented without the outlier rejecting scheme proposed in Sec. \ref{sec_match}. LIO-GVM(plane) replaces the proposed new residual metric with the point-to-plane distance \cite{xu2022fast}. As shown in Tab. \ref{table:odom}, both LIO-GVM(w/o s) and LIO-GVM(plane) turn worse compared to the original system for all the sequences, indicating that the proposed correspondence matching scheme and residual metric both contribute to the accuracy of the system. The accuracy of LIO-GVM(plane) tends to be much worse than LIO-GVM(w/o s), which is because the residual metric inherently reduces the weights for mismatches. The drastic accuracy degradation of LIO-GVM(plane) is because, without the proposed residual metric, all the adopted correspondences contribute equally to the estimation, even those with low similarity.

As illustrated in Tab. \ref{table:odom}, LIO-GVM achieves the best results for most sequences due to its ability to reject mismatched distributions (the proposed correspondence matching) and evaluate the variance divergence in the distribution covariance (the proposed residual metric). For \textbf{ntu\_*} sequences, FAST-LIO2 achieves comparable results with LIO-GVM, because the accurate IMU integration can approximately project source points to the correct feature in the target map. VoxelMap offers accurate pose estimation in some sequences but frequently diverges, especially for the \textbf{nc\_*} sequences equipped with a 64-line LiDAR. This divergence is attributed to its map update scheme. When the LiDAR remains stationary for several scans, distant voxels converge prematurely and stay undersampled. This issue is particularly obvious for ground voxels, leading to the majority of divergence occurrences along the z-axis. Conversely, LIO-GVM employs a continuously updated mapping scheme, ensuring that the performance remains stable for motion and LiDAR sampling rate variations.

The performance of all LIO systems is negatively impacted in \textbf{ntuo\_*} compared to \textbf{ntu\_*} sequences due to the equipment of a low-cost IMU. Nevertheless, LIO-GVM is able to mitigate the impact of IMU noises through its proposed correspondence matching scheme, which can reject outliers. Additionally, LIO-GVM incorporates a new residual metric that takes similarity into account during the estimation phase, allowing more effective correspondences to have a greater influence on the estimation. Consequently, LIO-GVM exhibits the best accuracy compared to other systems. However, all the methods have a common limitation in accommodating the pronounced changes in elevation in the \textbf{ntu\_*} sequences, as seen in Fig. \ref{fig:xyz} and Fig. \ref{fig:traj}. 

Overall, LIO-GVM has demonstrated the best performance regarding odometry accuracy, showing the effectiveness of the error-rejecting correspondence matching scheme and the distribution-based residual metric in achieving accurate pose estimation across various sensor setups, especially in long-term applications or with cheaper IMUs. 

\begin{table*}[!t]
\caption{evaluation of the temporal efficiency in sequences on mapping procedures}
\label{table:mapTime}
\centering
\resizebox{0.9\textwidth}{!}{
\begin{threeparttable}
    \begin{tabularx}{\textwidth}{>{\raggedright}X*{8}{>{\centering\arraybackslash}X}}
\toprule
\multicolumn{1}{c}{ } & \multicolumn{2}{c}{Gaussian Fitting} & \multicolumn{2}{c}{Correspondence Match}  & \multicolumn{2}{c}{Incremental Update} & \multicolumn{2}{c}{Total Map Time}  \\
\cmidrule(lr){2-3} \cmidrule(lr){4-5} \cmidrule(lr){6-7} \cmidrule(lr){8-9} 
 & ikd-Tree & GVM & ikd-Tree & GVM & ikd-Tree & GVM & ikd-Tree & GVM \\
\midrule
ntu\_02     & \textbf{0} & 11.61      & 25.05 & \textbf{1.29}      & \textbf{2.45} & 5.13   & 27.50 & \textbf{18.03} \\
ntu\_04     & \textbf{0} & 12.91      & 24.26 & \textbf{1.41}      & \textbf{3.96} & 5.34   & 28.22 & \textbf{19.66} \\
ntu\_10     & \textbf{0} & 11.86      & 24.27 & \textbf{1.34}      & \textbf{3.97} & 5.87   & 28.24 & \textbf{19.08} \\
ntu\_13     & \textbf{0} & 12.71      & 24.89 & \textbf{1.67}      & \textbf{3.62} & 5.18   & 28.51 & \textbf{19.56} \\
nc\_01  & \textbf{0} & 8.10       & 14.36 & \textbf{0.69}      & \textbf{0.78} & 2.97   & 15.14 & \textbf{11.76} \\
nc\_02  & \textbf{0} & 8.25       & 15.76 & \textbf{0.73}      & \textbf{0.92} & 2.84   & 16.68 & \textbf{11.82} \\
nc\_05  & \textbf{0} & 7.73       & 12.62 & \textbf{0.67}      & \textbf{0.57} & 2.84   & 13.19 & \textbf{11.24} \\
nc\_06  & \textbf{0} & 7.19       & 13.66 & \textbf{0.70}      & \textbf{0.81} & 3.23   & 14.47 & \textbf{11.12} \\
nc\_07  & \textbf{0} & 8.99       & 15.67 & \textbf{0.62}      & \textbf{1.11} & 3.18   & 16.78 & \textbf{12.79} \\
\bottomrule
\end{tabularx}
\begin{tablenotes}
    \footnotesize
        \item We evaluate the \textbf{average time consumed per scan [ms]}.
        \item \textbf{Bold} value stands for the best value.
\end{tablenotes}
\end{threeparttable}
}
\vspace{-7pt}
\end{table*}

\begin{figure}[!t]
\centering
    \footnotesize
    \subfloat[MCD\_VIRAL dataset.]{\includegraphics[width=0.34\textwidth]{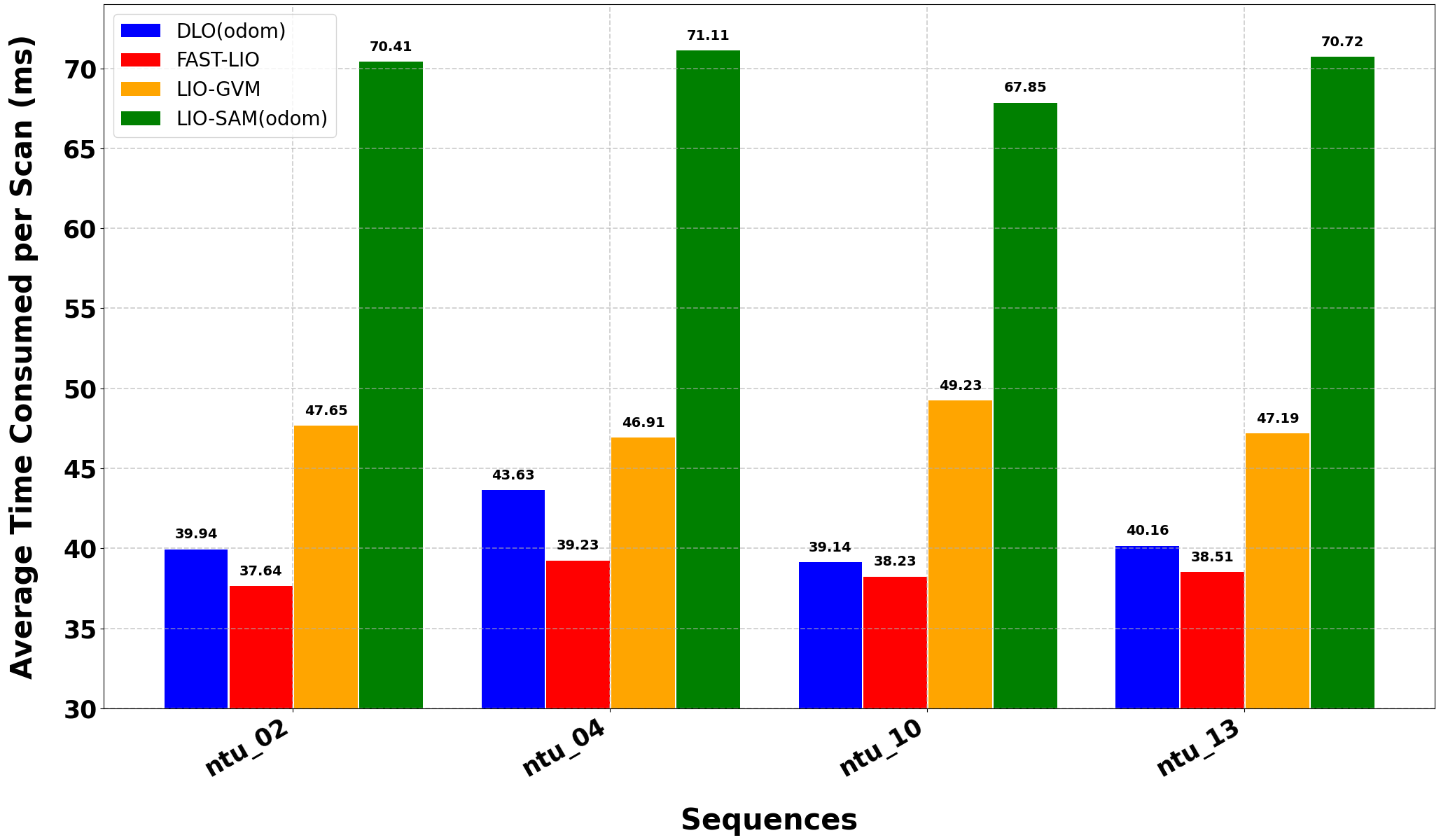} \label{fig1_a}}
    \quad 
    \subfloat[Newer College dataset.]{\includegraphics[width=0.34\textwidth]{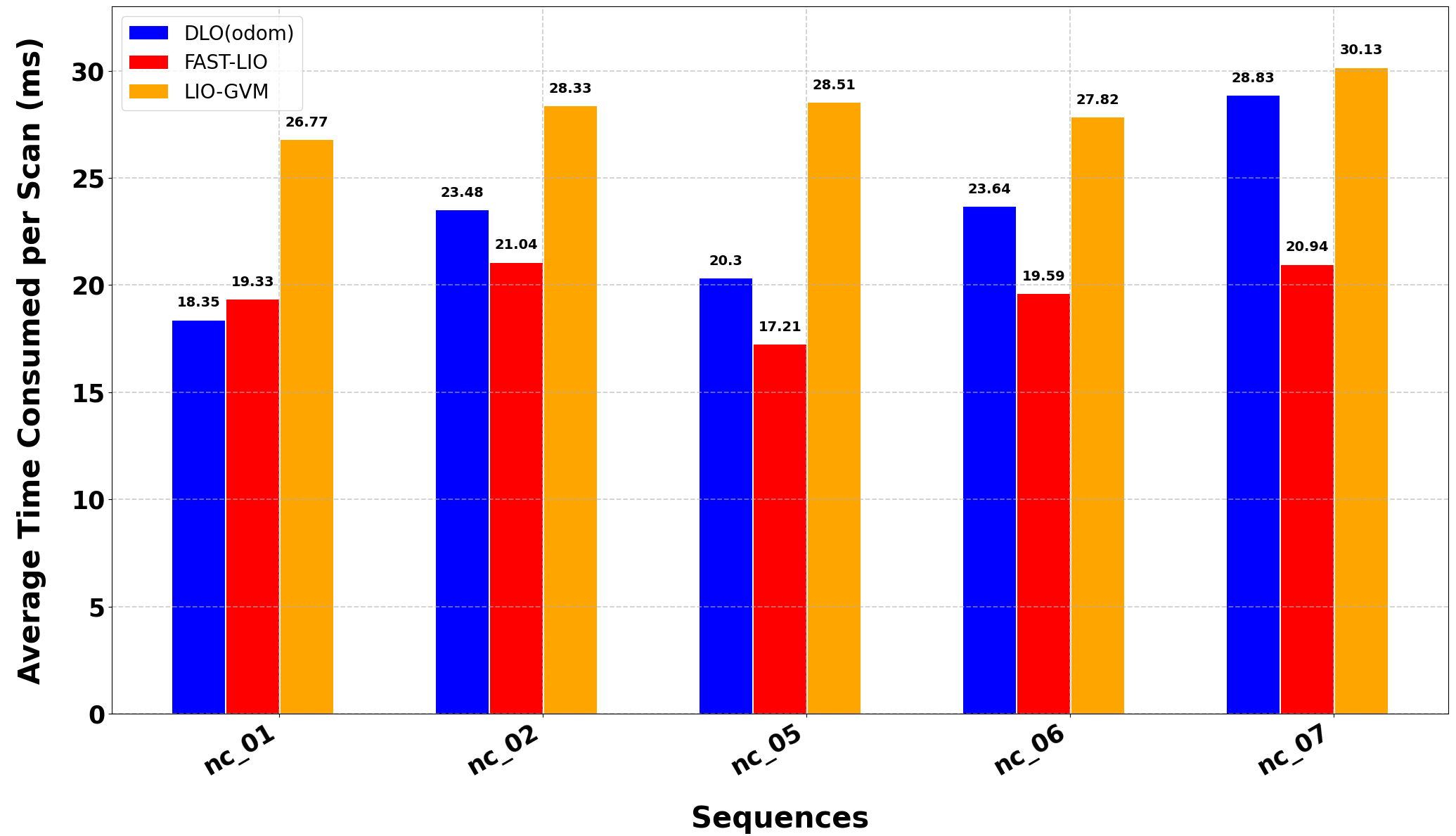}}
  \caption{The average time consumed per scan (ms) for all methods in different datasets.}
  \label{fig:avg_time}
  \vspace{-15pt}
\end{figure}

\vspace{-10pt}

\subsection{Evaluation of Storage Efficiency}
In this section, we evaluate the storage consumption of the global map generated by LIO-GVM. Specifically, in LIO-GVM, we have defined a custom point type using the Point Cloud Library (PCL). Each point encapsulates all the elements of $(\boldsymbol{\mu}_i, \mathbf{C}_i, M)$, comprising a total of four integer values and nine floating-point values. All the voxels are stored in a compressed binary .pcd file. For other systems, we configure their local map size to be 500m, ensuring coverage of all areas within the two datasets, and save all the points in the map to a compressed binary .pcd file. 

\begin{figure}[!t]
\centering
\includegraphics[width=2.5in]{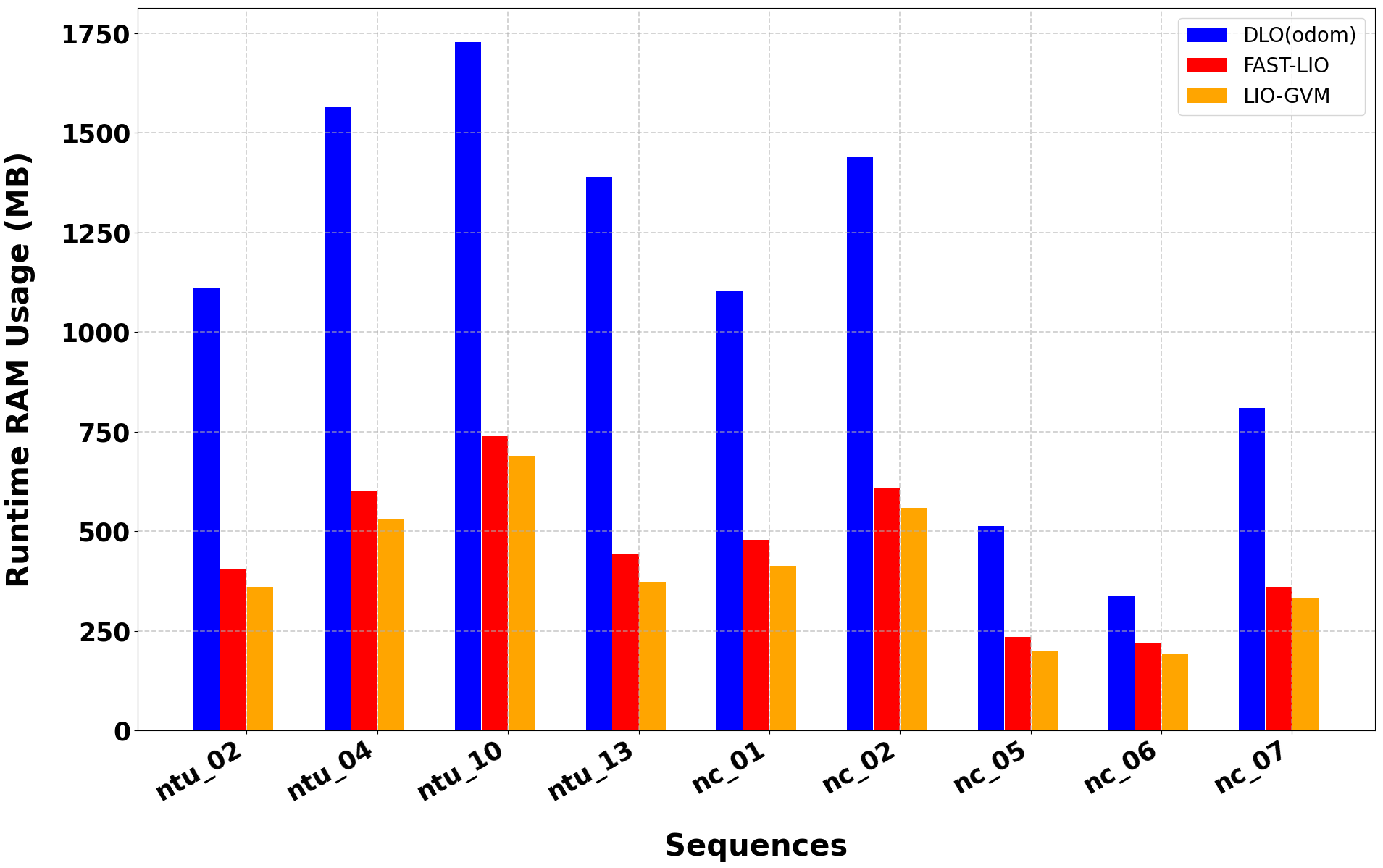}
\caption{The eventual runtime RAM usage for all the methods.}
\label{fig:ram}
\vspace{-10pt}
\end{figure}

As illustrated in Tab. \ref{table:mapStorage}, the map storage consumption of LIO-GVM outperforms other systems for all the sequences. This is because other systems need to ensure that the downsampling filter of the map is not excessively large (usually $<$ 0.5m) to preserve the geometric information of the points.  In contrast, our approach allows us to allocate a considerably larger voxel size (commonly $>$ 1m) compared to other systems while still offering a comprehensive representation of the intra-voxel point distribution. 

\begin{table*}[!t]
\centering
\caption{The constructed Map Storage analysis}
\label{table:mapStorage}
\resizebox{.9\textwidth}{!}{\begin{threeparttable}
\begin{tabularx}{\textwidth}{>{\raggedright}X*{9}{>{\centering\arraybackslash}X}}
\toprule
         & ntu\_02 & ntu\_04    & ntu\_10 & ntu\_13   & nc\_01 & nc\_02  & nc\_05  & nc\_06   & nc\_07 \\ 
\midrule                                                                                                                          
FAST-LIO2 & 6.07       & 21.83      & 19.13       & 22.19     & 7.93       & 9.24        & 1.15        & 0.65         & 5.62 \\
LIO-GVM     & \textbf{3.55}  & \textbf{12.61} & \textbf{11.39}  & \textbf{8.29} & \textbf{4.25}  & \textbf{4.60}   & \textbf{0.13}   & \textbf{0.48}    & \textbf{3.01} \\
\bottomrule
\end{tabularx}
\begin{tablenotes}
    \footnotesize
        \item The map storage is analyzed by evaluating the \textbf{storage consumption [MB]}. All the maps are saved as binary compressed PCD files.
        \item \textbf{Bold} value stands for the best value.
\end{tablenotes}
\end{threeparttable}
}
\end{table*}

\vspace{-10pt}

\subsection{Evaluation of Time Consumption}
In this section, we will evaluate the temporal efficiency of LIO-GVM, utilizing the same parameters as those deployed in our previous odometry evaluations for consistency. We first analyze the influence of the data structure on the mapping procedure, which involves correspondence matching and incremental updates. Given the superior performance of the ikd-Tree provided in \cite{xu2022fast} compared to other dynamic tree-based structures, our comparison will be limited to this structure. While the Gaussian fitting process does not directly participate in the mapping procedure, its time consumption is still accounted for in LIO-GVM for a fair comparison. We denote the data structure in LIO-GVM for the mapping procedure as GVM. The results are illustrated in Tab. \ref{table:mapTime}. GVM significantly outperforms ikd-Tree in correspondence matching because the time complexity of $k$-nearest neighbors searching for ikd-Tree is $\mathcal{O}(log(n))$, where $n$ is the tree size; but for GVM, managed in a hash table, the corresponding time complexity is $\mathcal{O}(1)$. ikd-Tree slightly performs better in the incremental update. Nonetheless, GVM demonstrates superior performance in total map time. 

In Fig. \ref{fig:avg_time}, we illustrate the average total time consumed per scan for all the sequences. The total time we evaluated includes all the procedures: data processing, pose estimation, and map update. Since LIO-SAM and DLO run map update in different thread with the pose estimation thread, we only evaluate their time consumed for state estimation. Notably, VoxelMap consumes an average of 141.3 ms per scan due to its demand for dense point cloud data. Even though the mapping procedure of LIO-GVM is faster than FAST-LIO2, FAST-LIO2 achieves the shortest total time for almost all the sequences. This is because LIO-GVM requires more geometric information from the input scan. For example, in \textbf{ntu\_*} sequences, FAST-LIO2 employs fewer feature points than LIO-GVM (4638 versus 5741). Consequently, LIO-GVM spends more time on pose estimation. However, as illustrated in Tab. \ref{fig:ram}, the fast performance of FAST-LIO2 and DLO comes with the trade-off of increased runtime RAM usage, while LIO-GVM can achieve comparable temporal efficiency with lower RAM usage. 

\vspace{-7pt}

\section{Conclusion}
In conclusion, this letter proposes an accurate filter-based LIO system denoted as LIO-GVM, which models the points as Gaussian distributions for robust correspondence matching. The key innovation of the system is the proposed new residual metric. Unlike previous methods that solely measure distance, our approach incorporates variance disparity, thereby enhancing the system's performance. Benefiting from the new metric, the map of LIO-GVM can be maintained in a simple yet efficient manner. We demonstrate the reliability and efficiency of LIO-GVM through extensive experiments on different datasets. The limitation of LIO-GVM is that it requires equipment with higher resolution LiDAR to ensure informative Gaussian fitting of the input scan. Additionally, the voxel size is not adaptive to the environment. To address these limitations, future work will focus on developing an adaptive voxel representation scheme for the global map. 

\vspace{1ex} 

{\appendix
\subsection{Supplemental Material} \label{sec_apd_a}
Please find details in the supplemental material: \href{https://github.com/Ji1Xingyu/lio_gvm/blob/main/supplemental_material.pdf}{supplemental\_material.pdf}. 

\subsection{Derivation of the Proposed Residual Metric} \label{sec_res_proof}
Let's denote $\left( {}^{G}\hat{\boldsymbol{\mu}}_j^n - {}^G\boldsymbol{\mu}_i \right)$ as ${}^G \Tilde{\boldsymbol{\mu}}$, then equate \eqref{eq_weighted_md} with the second term in MAP problem, we'll have: 
\begin{equation} \label{eq_ob}
\| \mathbf{z}_k^n\|^2_{\mathbf{V}_k^{-1}} = \sum_{j=0}^m (s_{j,i}^n)^2 {}^G \Tilde{\boldsymbol{\mu}}^T (\hat{\mathbf{C}}_j^n + \mathbf{C}_i + \alpha \mathbf{I})^{-1} {}^G\Tilde{\boldsymbol{\mu}}.
\end{equation}
To avoid the singularity issue, we first perform eigendecomposition on $(\hat{\mathbf{C}}_j^n + \mathbf{C}_i + \alpha \mathbf{I})^{-1}$ to have $(\hat{\mathbf{C}}_j^n + \mathbf{C}_i + \alpha \mathbf{I})^{-1} = \left(\mathbf{U}_j^n\right)^T \left(\boldsymbol{\Lambda}_j^n\right)^{-1} \mathbf{U}_j^n$. Subsequently, the eigenvalues are normalized and clamped at $10^{-4}$: $\lambda_{t,j}^{n} = max(\lambda_{t,j}^{n} / trace({\Lambda}_j^n), 10^{-4}), t = 1, 2, 3$. In order to obtain the same formula as the left-hand side, we perform matrix decomposition to have $\left( {\Lambda}_j^n \right)^{-1} = \left(\mathbf{Q}_j^n\right)^T  \mathbf{V}_k^{-1} \mathbf{Q}_j^n$, which is quite easy since $ \left( {\Lambda}_j^n \right)^{-1}$ is a diagonal matrix. Thus, 
\begin{equation}
\begin{aligned}
\eqref{eq_ob} &= \sum_{j=0}^m (s_{j,i}^n)^2 {}^G \Tilde{\boldsymbol{\mu}}^T (\hat{\mathbf{C}}_j^n + \mathbf{C}_i + \alpha \mathbf{I})^{-1} {}^G \Tilde{\boldsymbol{\mu}} \\
&= \sum_{j=0}^m s_{j,i}^n {}^G \Tilde{\boldsymbol{\mu}}^T \left(\mathbf{U}_j^n\right)^T \left(\mathbf{Q}_j^n\right)^T  \mathbf{V}_k^{-1} s_{j,i}^n \mathbf{Q}_j^n \mathbf{U}_j^n  {}^G \Tilde{\boldsymbol{\mu}} \\
&= \sum_{j=0}^m (\mathbf{z}_{k,j}^n)^T \mathbf{V}_k^{-1} \mathbf{z}_{k,j}^n,
\end{aligned}
\end{equation}

where:
\begin{equation}
    \begin{aligned}
        \mathbf{z}_{k,j}^n &= s_{j,i}^n\mathbf{D}_j^n \left( {}^{G}\hat{\boldsymbol{\mu}}_j^n - {}^G\boldsymbol{\mu}_i \right), \\
        \mathbf{D}_j^n &= \mathbf{Q}_j^n\mathbf{U}_j^n.
    \end{aligned}
\end{equation}

\bibliographystyle{IEEEtran}
\bibliography{mybib}
\end{document}